\documentclass[runningheads]{llncs}
\usepackage{graphicx}
\usepackage{hyperref}
\usepackage{color}

\begin{document}
\title{Detection and Pose Estimation of flat, Texture-less Industry Objects on HoloLens using synthetic Training}
\titlerunning{Augmented Reality-assisted sorting of Industry Objects}

\author{Thomas Pöllabauer\inst{1,2}\orcidID{0000-0003-0075-1181} \and \\
Fabian Rücker\inst{1}\orcidID{0000-0003-4071-8642} \and \\
Andreas Franek\inst{1}\orcidID{0009-0009-6823-8414} \and \\
Felix Gorschlüter\inst{1}\orcidID{0000-0003-2897-1899}}

\authorrunning{T. Pöllabauer et al.}

\institute{Fraunhofer Institute for Computer Graphics Research IGD, Germany \and
Interactive Graphics Research Group, TU Darmstadt, Germany
%\email{thomas.poellabauer@igd.fraunhofer.de}%\\
\email{\{thomas.poellabauer,fabian.ruecker,andreas.franek,felix.gorschlueter\}\\@igd.fraunhofer.de}
}

\maketitle              % typeset the header of the contribution
\begin{abstract}
Current state-of-the-art 6d pose estimation is too compute intensive to be deployed on edge devices, such as Microsoft HoloLens (2) or Apple iPad, both used for an increasing number of augmented reality applications. The quality of AR is greatly dependent on its capabilities to detect and overlay geometry within the scene. We propose a synthetically trained client-server-based augmented reality application, demonstrating state-of-the-art object pose estimation of metallic and texture-less industry objects on edge devices. Synthetic data enables training without real photographs, i.e. for yet-to-be-manufactured objects. Our qualitative evaluation on an AR-assisted sorting task, and quantitative evaluation on both renderings, as well as real-world data recorded on HoloLens 2, sheds light on its real-world applicability.

\end{abstract}

\section{Introduction}
Supporting manual industrial sorting tasks digitally is an important challenge for the digitalization of industrial processes. Detecting relevant objects is difficult even in use cases with re-occurring shapes, such as in retail, but even more so when providing on-demand custom products with potentially only a single instance being produced. One part of the problem is lacking automation: with a potentially endless amount of shape variations, labelling manually becomes next to infeasible. Another problem is the inability to train detectors and classifiers of the yet-to-be-manufactured objects, i.e. without any real-world (and correctly annotated) image data. Finally, one has to make this tool accessible and easy to use for productive work. Augmented reality, the visual enrichment of the real-world, allows to elegantly deliver this assistance, but is most often used on hand-held devices, such as smartphones, which interferes with the sorting-by-hand process.\\
This work uses single-view RGB-only pose estimation to propose a system for user guidance in a productive environment. We address all of the problems listed above: We automatically extract all required information from the available manufacturing documents, requiring no manual oversight. Using this data, we derive all necessary information to train an object detector and a state-of-the-art 6D (rotation + translation) object pose estimator. Lastly, we demonstrate a graphical user interface on a head mounted display (HMD) or tablet to support the sorting of objects, the user has not yet seen. \\
Our contribution is the qualitative and quantitative demonstration of current deep learning algorithms on flat, texture-less objects requiring no real-world data, only two-dimensional vector representations taken from the manufacturing documents. We explicitly shed light on the applicability on real-world industry objects, implementing an augmented reality assistance application for object sorting. All of our training is conducted using only synthetic data, i.e. renderings of our automatically constructed meshes. By adopting a client-server solution, we stream the video images from the device to a backbone for computation, alleviating the computation constraints and bringing workstation-level performance to low-power edge devices. Finally, our approach is highly modular, allowing to swap individual components to better suit specific problem domains. \\

This work is structured as follows: First we give an overview on the relevant literature on AR, object pose estimation, and the major problem when training with synthetic data, domain gap. Next we present our approach, followed by qualitative, as well as quantitative results. Finally we discuss some limitations and future work.

\section{Related Work}

We will introduce relevant work in augmented reality applications using head mounted displays, as well as work tackling the detection and pose estimation problem. Finally, we present relevant work dealing with the domain gap between real-world and synthetic images. 

\subsection{AR and HMD}
Augmented Reality is a technology that extends the perceived reality digitally \cite{ARintro} e.g. by adding contextual information or highlighting points of interest. When using a handheld device (e.g. a tablet or smartphone), this is achieved by rendering digital content on top of the live camera feed \cite{smartphoneAR}. Therefore the digital overlay is only visible when the user looks at the device and the respective object is captured by the device's camera.\\
Alternatively Head-Mounted-Displays (HMD's) can be used for AR applications. These augment the environment directly within the user's field of view, either by modifying a displayed camera live-feed (video see-through), or by superimposing digital content over the environment that is perceived through transparent lenses (optical see-through) \cite{vidoptst}. Common AR-HMD's are for example Microsoft HoloLens (2), Magic Leap or Nreal Light. All mentioned devices offer an optical see-through experience with a restricted field of view for digital content (30-52° diagonal FoV). Because the device does not need to be held, the user's hands are free for other tasks.\\
Besides entertainment use-cases (e.g. displaying contextual information during sports events), augmented reality is increasingly relevant for industrial use-cases \cite{IARsurvey} namely quality assurance, manufacturing and assembly procedures \cite{assembly}. 
Since neural networks require high computing performance, current state-of-the-art pose estimators cannot run locally on an AR-HMD. Instead the device is used for visualization only, receiving and displaying the results from the AI-cloud. \cite{hololensTrack} uses such an approach to solve object detection (2D), but not 6D pose estimation, like we do. To the best of our knowledge the closest work to ours is \cite{competition}: They present 6D detection and tracking for similar and non-textured objects trained on synthetic data. The main differences are: 

\begin{itemize}
  \item Ours is a single/one-shot pipeline, easily adapted for
specific use cases, while they specifically extend PVNet and rely on
multiple images for refinement.
  \item We use real industry objects and
can deal with rotationally symmetric objects, which they specifically say, should be avoided.
  \item Our approach can work with identical objects of different color, while they exclusively use gray-scale
images, discarding most color information.
  \item They rely on object geometry edges for refinement, while we could use realistic
textures, if available.
  \item Finally, we illustrate our solution on the relevant hardware and evaluate on real-world HoloLens 2 images.
\end{itemize}

% Ours is a single/one-shot pipeline, easily adapted for specific use cases, while they specifically extend PVNet and rely on multiple images for refinement. We use real industry objects and can deal with rotationally symmetric objects, which they specifically say, should be avoided. Our approach can work with identical objects of different color, while they exclusively use gray-scale images, discarding most color information. Also, they rely on object geometry edges for refinement, while we could use realistic textures, if available. Finally, we illustrate our solution on the relevant hardware and evaluate on real-world HoloLens 2 images. 
In summary, our pipeline is a more general approach on how state-of-the-art 6D pose estimation can be made available on edge devices.
% They distinguish between close- and far-range mode 

\subsection{Object Pose Estimation}
Estimating the 6D pose of an object from an image is one of the major challenges for augmented reality applications. In order to accurately render additional information on top of the object in the scene, one requires a low degree of error due to the negative influence on user experience. Among the solutions for this problem the most prominent in recent literature rely heavily on deep learning. A good overview across different state-of-the-art approaches is presented in the BOP Challenge \cite{bopchallenge2020}.
Among the different approaches the most important differences, aside of the distinction between non deep learning and deep learning-based methods, are in the modalities used (i.e. RGB-only or additional information such as depth), single- \cite{gdrnet,sopose,deepim} versus multi-stage \cite{singleshotpose,epos,pix2pose,pvnet,dpod,cosypose,hybridpose}, monocular versus stereo \cite{keypose,stereobj-1m}, template-based approaches \cite{linemod,hodan_template} versus local feature-based methods \cite{singleshotpose,epos,pix2pose,pvnet,dpod,deepim,cosypose,hybridpose,gdrnet,sopose}, and, with the rise of differentiable rendering, as well as neural rendering, analysis-by-synthesis approaches, such as NeRF-based approaches \cite{inerf}. Also, the rise of the transformer model in the domain of neural language processing begins to find its way into vision as well, resulting in the adaption of transformer architectures in applications such as image classification, object detection, segmentation, image inpainting, and image generation \cite{igpt,swin,glide}.
Since our application was to be run on HoloLens, we were limited by its sensors, restricting us to RGB-only methods. Also, we wanted to be able to work with a single image, leading us to focus on single-view or single-shot approaches. 
Among the suitable approaches SingleShot\cite{singleshotpose} is arguably the simplest, extending the YOLOv2 object detector to, instead of predicting only the 2D bounding box, predict the projections of the 8 3D bounding box corners and the centroid of the object, before applying Perspective-n-Point (PnP). Pix2Pose \cite{pix2pose} similarly predicts pixel-wise 3D coordinates and estimates the pose via PnP and RANSAC. EPOS \cite{epos} again uses PnP+RANSAC, but predicts correspondences between pixels and object surface fragments. HybridPose \cite{hybridpose} is a multi-step pipeline, first predicting keypoints, edge vectors, and symmetry correspondences, before making a coarse prediction, followed by a refinement step. All previous three approaches leverage encoder-decoder networks. PVNet \cite{pvnet} introduces a voting scheme, requiring pixels to vote for keypoint locations, making the predictions more robust to occlusion and truncation. It again, uses a PnP solver. DPOD \cite{dpod} predicts a mask and 2D-3D correspondences. Also, they compare training on synthetic and real data. GDR-Net \cite{gdrnet}, similarly to EPOS, leverages surface regions as well as 2D-3D correspondences, but replaces the conventional PnP solver with a learnable Patch-PnP alternative, coming up with an end-to-end trainable estimator. SO-Pose\cite{sopose} predicts 2D-3D correspondences plus an object mask, as well as self-occlusion maps, again in an end-to-end trainable estimator. DeepIM \cite{deepim} uses an iterative matching procedure, estimating the delta pose between the rendering of an initial pose estimate with the target view. CosyPose \cite{cosypose} builds upon DeepIM by reformulating the loss function and using the iterative matching process as phase one out of three. The output of phase 1 are object candidates per image. CosyPose, being a multi-view solution, uses these candidates in phase 2 and 3 to first, match objects across views, before finally optimizing to get a global solution for the scene. \cite{coupled} again uses an iterative refinement strategy, but introduces end-to-end differentiability optimizing both correspondences as well as pose estimates. 

\subsection{Domain Gap}
Synthetically producing the data required to train ever larger networks attracted a lot of interest. In computer vision, the idea to use rendering, the well established science of how to produce two-dimensional images of 3D scenes, is an obvious solution to train models in domains with little or no real photographs. Doing so introduces a new problem called domain gap: the difference between a photograph of a real 3D scene and the rendering of digital 3D geometry. Sources of domain gap include unrealistic materials, physically incorrect lighting transport with rasterization pipelines, the placement of objects, etc. Relevant work, dealing with or circumventing the domain gap problem, includes \cite{pretraining}\cite{hinterstoisser_annotation_saved}\cite{bridgingWithSyntheticData}\cite{dope}\cite{reversal}\cite{deception}\cite{cornet}. The first work introduces the idea of freezing a network pre-trained on real images, the second highlights (among other contributions) the importance of meaningful 3D-scenes, i.e. foreground and background depict the same lighting, the third and fourth deal with data set composition and combining non-realistic and realistic images, the fifth and sixth propose the use of gradient reversal for active domain adaptation (the bridging between two domains) during training, and the last one uses features shared between synthetic and real data, i.e. object corners. Finally, \cite{self6d} proposes to circumvent the problem using unsupervised learning, an approach, we cannot adopt for lack of any real images prior to manufacturing. 

\section{Approach}
\begin{figure}[htb]
  \centering
    \includegraphics[width=0.8\textwidth]{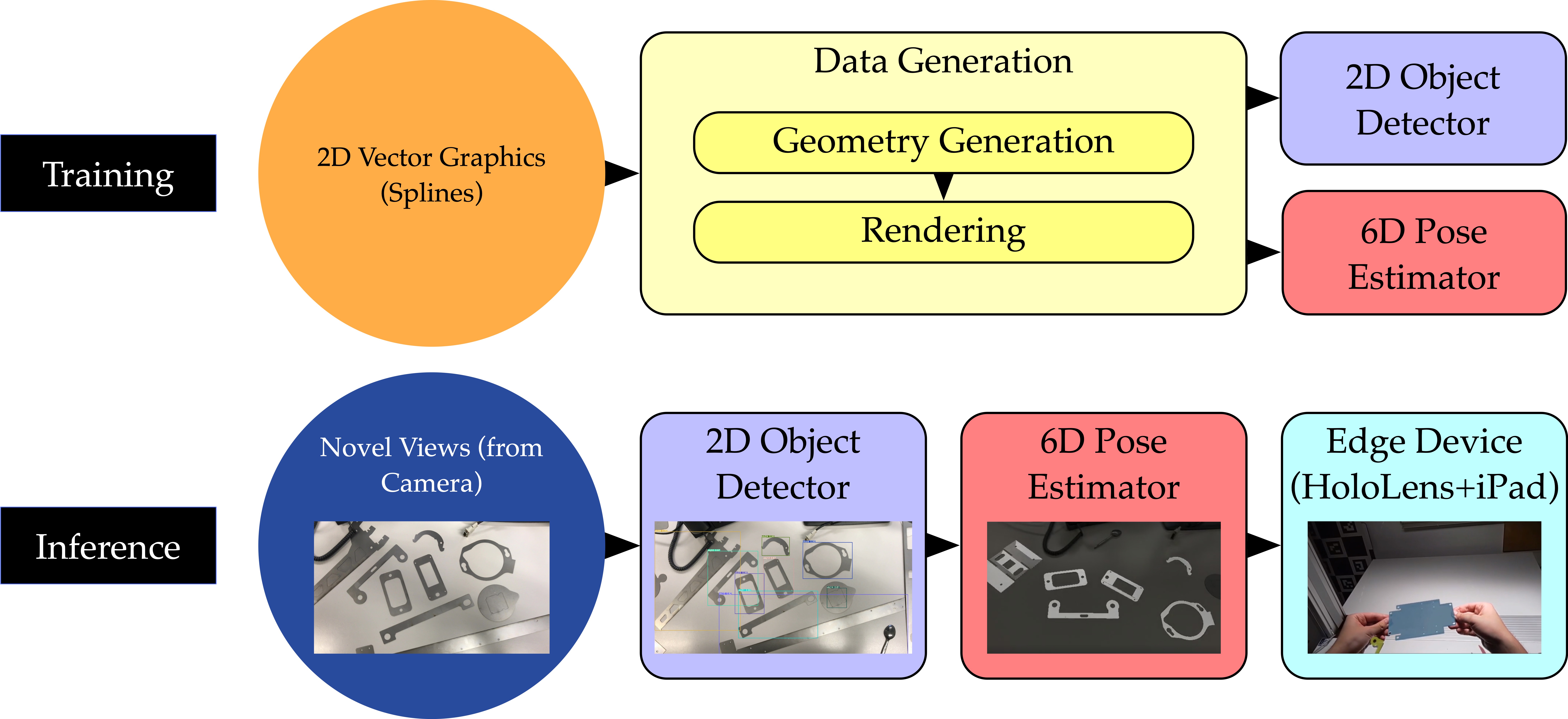}
  \caption{\label{fig:approach}
           Functional blocks in our approach. First we extract the shapes (curves) from the manufacturing documents (Splines). Next, we create meshes from the extracted shapes (Geometry Generation) and use them to create our physically-based, photo-realistic training dataset, as well as our non-photorealistic dataset (Rendering). Combining both datasets, we train our object detector and pose estimation pipeline. At inference, given a real-world camera stream, we process the data image-per-image and get per image detections and 6D pose vectors, which are displayed either on HoloLens 2 or iPad.}
\end{figure}
 Our goal is to detect yet-to-be-manufactured objects and estimate their pose. In addition, we can only rely on the manufacturing documents, which are fed to the machine that is cutting the objects from a metal sheet. These schematics contain the patterns in form of splines.
%Since our use case involves the detection of a great number of different objects in a very short amount of time, we limit the available training time to roughly 7 days per object set on a single high-performance GPU (Nvidia RTX 3090). 

Our solution (Figure \ref{fig:approach}) consists of five main stages: First, process the manufacturing documents to derive data for training (\ref{data_generation_blender}). Second, generate training data for object detection and object pose estimation (\ref{data_generation_training}). Third, selecting our network architectures and algorithms (\ref{architecture_selection}), and fourth, training of our networks (\ref{training_details}). Finally, our frontend application, running on the edge device (\ref{on_device_application}).

\begin{figure}[b]
  \centering
    \includegraphics[width=1\textwidth]{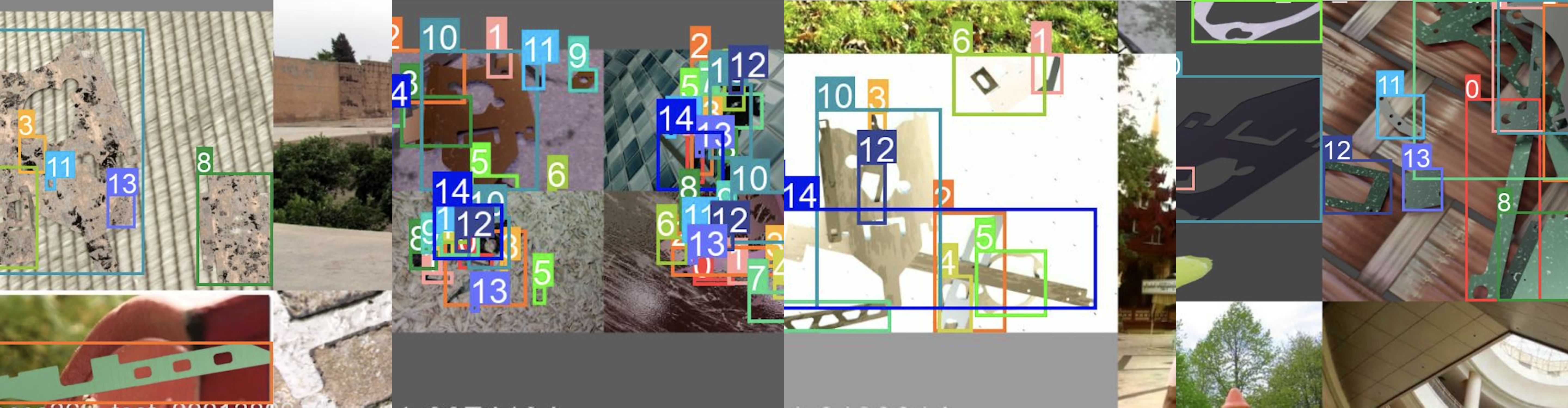}
  \caption{\label{fig:synth_samples}
           Samples of our synthetic training data. Note that we make very few assumptions about material or lighting: we randomly choose the coefficients for specular and diffuse reflection, apply random textures, and a random number of (randomly placed) light sources. Also, one sees the effect of mosaic augmentation (as proposed in \cite{yolov4}).}
\end{figure}

\subsection{Data Generation}
\subsubsection{Target Object Description}
Our objects are a representative sample of shapes produced at a leading industry supplier. The shapes are cut from a 1mm strong metal sheet. For our application we assumed the objects to be rigid, an assumption that proved to be appropriate for all but the longest objects (08, 15 as depicted in Table \ref{fig:ObjectDetectionResults}) when held in hand. Also we chose not to use the fact, that the objects can be considered 2-dimensional, so that our approach can be adopted for more general 3-dimensional objects. 

\subsubsection{Manufacturing Documents to 3D Models}
\label{data_generation_blender}
In order to train our image processing pipeline, we first have to create geometry based on the provided data. The manufacturing documents are available in an xml file-format, which contains SVG descriptions of the objects to be detected, together with their respective object category. A data preparation pipeline imports the xml-file, exports the different objects as individual SVG files, and produces meshes via the 3D modeling software Blender (\url{www.blender.org}) by adding the metal sheet's thickness to the base shape. The 3D models are used to generate our training data in the next step.

\subsubsection{Training Data Generation}
\label{data_generation_training}
We used two rendering solutions, with which we rendered two sets of images comprising our training data. This approach was inspired by work done at Nvidia \cite{dope}. First, we render non-photorealistic images on top of random real world photographs. Second, we add a second set of photo-realistic images together with physics-based object placement, produced by Blender's raytracing solution Cycles. The first set of non-photorealistic images is produced using the render solution deployed by CosyPose (pybullet). For our photo-realistic set we use the open source BlenderProc solution \cite{blenderproc} to generate BOP compatible datasets \cite{bopchallenge2020}. Samples of our training set are shown in Figure \ref{fig:synth_samples}. 

\subsection{Architecture Selection}
\label{architecture_selection}
For our object detection, we took a closer look at Mask R-CNN \cite{mask-rcnn}, as well as YOLOv3 \cite{yolov3}, YOLOv4 \cite{yolov4}, and YOLOv5 (\url{https://github.com/ultralytics/yolov5}). We decided on YOLOv5, since it reported highest performance, as also stated by \cite{yolo_comparison}. The detector can, however, easily be swapped for an existing architecture or some other algorithm, such as Scaled Yolov4 \cite{scaled-yolov4} and/or EfficientDet \cite{efficientdet}) or one of the many new flavors of YOLO, such as YOLOX \cite{yolox}, YOLOR (\url{https://viso.ai/deep-learning/yolor/}), or PP-YOLO \cite{ppyolo}.\\
As for pose estimation, we considered a range of different algorithms, such as SingleShot \cite{singleshotpose}, Pix2Pose \cite{pix2pose}, HybridPose \cite{hybridpose}, EfficientPose \cite{efficientpose}, EPOS \cite{epos}, and CosyPose \cite{cosypose}. Tested on individual objects, SingleShot failed to produce usable results, while EfficientPose struggled with occlusion and rotation. We dropped HybridPose for its complexity. Among the remaining approaches, we preferred CosyPose mainly for four reasons: its modularity, which allows to easily swap out the detector (unlike EPOS), the better ratio between speed and prediction quality, as well as the possibility to combine multiple views for estimation, in case we wanted to do so at a later time. Finally, it outperformed the other approaches in \cite{bopchallenge2020}.

\subsection{Training Details}
\label{training_details}
%Since we wanted to be able to train all models within one week on a single GPU, we split the training time into 4 days for YOLO, 1.5 days for CosyPose's coarse, and 2 more days for CosyPose's refiner network.\\
We train YOLOv5 in its largest configuration (X) on 28.401 images containing target objects and add 30.425 images without. The high amount of "empty" images greatly reduces the false positive rate, which would lead to wrong detections popping up, which is very distracting when using an AR application. Our training set of 28.401 images consists of roughly 22.000 physically-based renderings (rendered with Cycles) and 7.000 images with objects cropped onto random backgrounds. \\
%We trained three models, the difference being the time available for training: one model trained for half a day (0.5D), one trained for 4 days (4D), and the final one for 500 epochs, or almost 26 days (26D) on a single GPU. 
We varied the image size from 512x512 to 1536x1536 pixels during training to increase scale invariance, used a batch size of 8, and SGD for optimization. As for augmentation, aside from our background replacement within the training data (masking the object and using random natural images as background), we only used the defaults of YOLOv5, such as mosaic augmentation (desribed in \cite{yolov4}). For initialization we resorted to using weights pre-trained on COCO. \\
For CosyPose, we experimented with different sizes of EfficientNet (B0, B3, B6), but in the end went with B3 as used in the original paper making a comparison of results more meaningful. 
%This, together with YOLO used for detection, leads to a speed up at inference of roughly 35\% (4,03 fps to 5,47 fps @ Nvidia RTX 3090) compared to the original Mask-RCNN + B3 configuration. However, since inference time is of secondary concern once one uses tracking (see Future Work \ref{future_work}) and the training time to performance ratio didn't improve to a noteworthy degree, we continued using the B3 configuration. 
We generated 250.000 images using the provided rendering script, and added these to our raytraced images used to train YOLO. Then we trained for 270 (coarse) / 350 (refiner) epochs, rescaling the input to 960x960 pixels. Parameter-wise we mostly relied on the original paper's values, only scaling the learning rate warm-up and the learning rate decay phases. Also, these networks were trained from scratch.

\subsection{Sorting App}
\label{mobile_app}
The sorting application facilitates the manual sorting process by guiding the user with color-cues. Different colors represent different object categories and each object is overlayed with the color of their respective category. An illustration is given in Figure \ref{fig:quali_hololens}. The overlay of the object is the respective colored 3D model, which is placed based on the pose estimation of the neural network and anchored to the HoloLens's coordinate system. Anchoring the object reduces the need for object tracking as long as the real object stays in place. Nevertheless, since the objects are moved around and picked up during the sorting process, the pose estimation needs to be updated regularly and cannot be reduced to finding the object on a plane table.
We chose a parameterization to achieve roughly 5 pose updates per second, which we find is a good compromise between the quality of the pose estimates and responsiveness on the front end. 
All in all the complexity of the sorting task, from the user perspective, is reduced to sorting scattered objects by color, which is much easier for human beings than sorting purely based on geometry.\\
To make the choice of the final display device independent of the computation requirements, we use a server-client architecture. All heavy computing (detection + pose estimation) is done on the server side, whereas the end user device (in our case HoloLens 2 or iPad) only sends images, receives and renders results. This approach enables deploying the solution to a wide variety of devices, as well as to incorporate it in a wider service architecture.\\
Currently both - our HoloLens and our iPad application - use the Unity Game Engine (\url{www.unity.com}) for providing the frontend. %Having a single code base allows to easily add features to all devices at once.

% Details on the communication part are requested:  It would be interesting to include some technical details about the communication with the visualization device (e.g. which codec is used for streaming the camera image, are special libraries used for streaming the video data of the hololens with unity, how long is the total delay introduced by streaming?)

\section{Qualitative Results}
\begin{figure*}[tbp]
  \centering
  \includegraphics[width=0.98\textwidth]{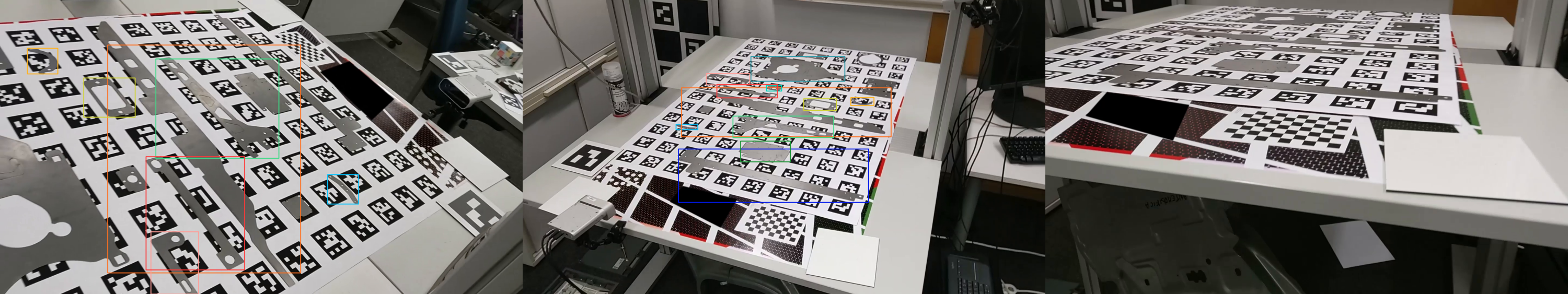}
  \caption{\label{fig:quali_detections}%
            Impression of our detection results. We achieve a high recall and good classification (see Figure \ref{fig:detection_graphs} for quantified results). We note 2 problems: First, with very flat viewing points performance drops drastically (rightmost image). Second, some camera movements lead to strong blurring and to the detector finding next to (or absolutely) nothing. \href{https://www.dropbox.com/s/03gvn69vdrs53zs/Qualitative\%20detection\%20results2.mp4?dl=0}{Please watch the video version for illustration}. 
          }
\end{figure*}

%\subsection{On-Device Application}
\label{on_device_application}
\begin{figure*}%[tbp]
  \centering
  \includegraphics[width=1.0\textwidth]{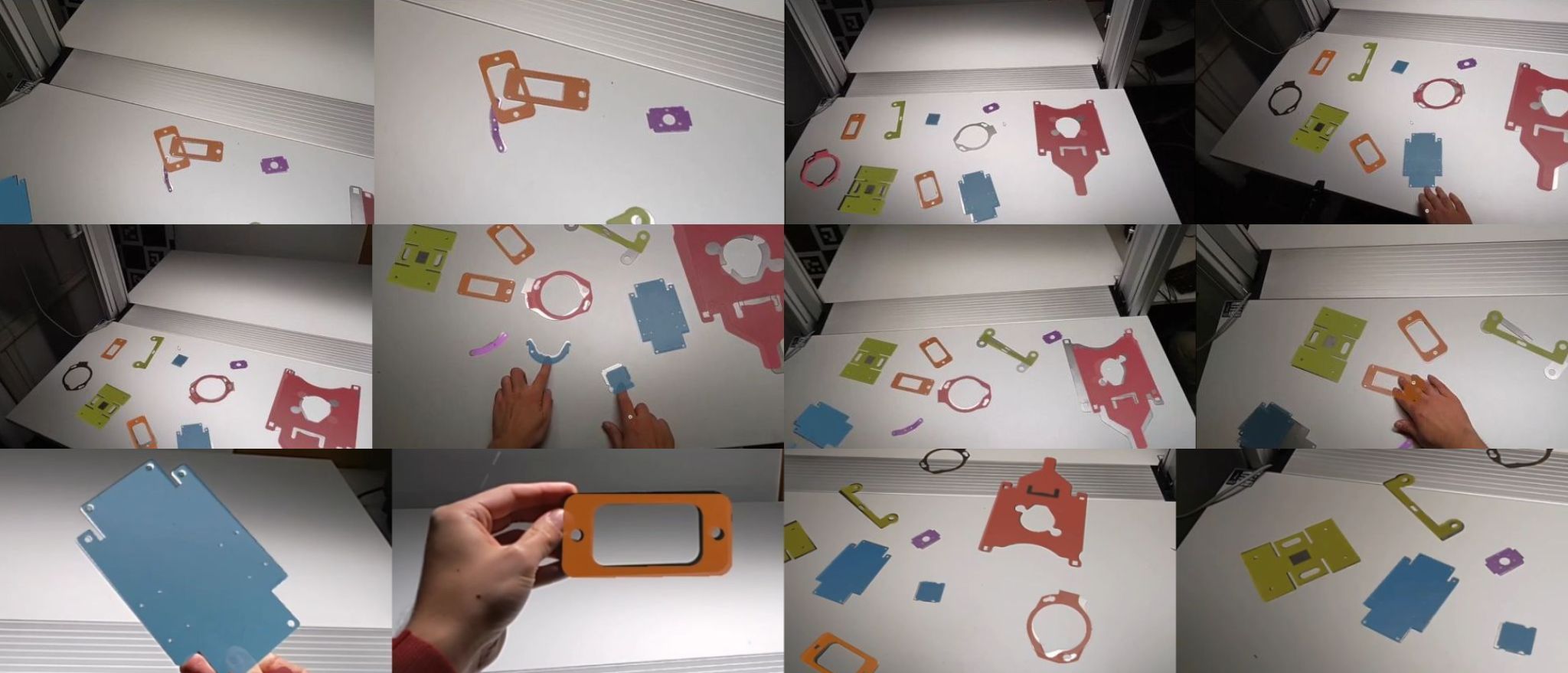}
  \hfill
  \caption{\label{fig:quali_hololens}%
            Qualitative results of our end-to-end results on HoloLens 2. Note that we do not use tracking at this point, only single-shot results. We do, however, place the colored objects with regards to the real world coordinate system, which makes them stay in place when the user moves. This makes our solution "real-time capable" although only 3-4 images are processed in the backend. 
  \href{https://www.dropbox.com/s/gasisytmtuvi1sa/TeaserARSorting_v2.mp4?dl=0}{Please watch the video version}. 
          }
\end{figure*}
\noindent The AR application seeks to provide an easy and intuitive sorting experience, where the user only needs to group the objects by color. Our provided videos show the functionality and user experience.\\
% Man sieht, man kann damit arbeiten. Die Intention kann erfüllt werden (Handsfree absortieren)
% Unterschied der Bewegung des Ipads und der HoloLens
%We present results both for HoloLens 2 and the iPad app, for several reasons: First and foremost the built-in cameras differ, which has an effect on the image quality and therefore detection quality. Secondly the expected camera movement is different. While a tablet is usually moved consciously, because it is held in hand and its camera needs to be pointed in the direction of the point of interest, a HMD like HoloLens 2 is highly influenced by unwitting and fast head-movements. Lastly the sorting procedure itself differs since using the tablet occupies a hand whereas using the HMD facilitates a hands-free sorting experience, which speeds up the sorting process.\\
%The iPad application offers the same functionality, but the sorting process is arguably slower since the user needs to either keep the tablet in one hand in order to see the color overlay or memorize a few objects' colors before sorting them, which is prone to error.
We present qualitative results from the detection network on HoloLens 2 video data in Figure \ref{fig:quali_detections} and from our end-to-end application in Figure \ref{fig:quali_hololens}. Please note that the latter results depict the real user experience, i.e. including latency caused by sending and receiving results to and from the backbone. To get a better impression, please have a look at the provided \href{https://www.dropbox.com/s/gasisytmtuvi1sa/TeaserARSorting_v2.mp4?dl=0}{sample video}.\\
We see strong detection performance both for predicting 2D bounding boxes, as well as end-to-end. Problems arise with fast camera movements, in which the HoloLens camera introduces blur, as well as with very flat viewing angles (failure case: Figure \ref{fig:quali_detections}, bottom right). End-to-end the pose estimates are overall sufficiently accurate for assistance purposes. The main issue is that, because of latency, the overlays are positioned inaccurately, while the objects are being moved.

\begin{table}%[t]
\centering
\caption{\label{tab:detector_model_size_comparison}
           Effect of detector model size and number of training epochs on performance on real test data. Model M seems to reach a saturation point somewhere between 1000 and 2500 epochs, while model L and X further improve the results. Performance is measured in mean average precision (mAP), higher is better.}
\begin{tabular}{ |p{1.4cm}|p{1.4cm}|p{1.3cm}|p{1.3cm}|p{1.2cm}| }
	\hline
	\multicolumn{5}{|c|}{Effect of Model Size (mAP@0.5 $\uparrow$)} \\
	\hline
	M @ 1000 & M @ 2500 & L @ 1000 & L @ 2500 & X @ 500  \\
	\hline
	 0.615 & 0.625 & 0.669 & 0.693 & \textbf{0.758}  \\
	\hline
\end{tabular}
\end{table}

% \begin{table}%[t]
% \centering
% \begin{tabular}{ |p{0.8cm}|p{0.8cm}|p{0.8cm}|p{0.8cm}|p{0.8cm}|p{0.8cm}|p{0.8cm}|p{0.8cm}|p{0.8cm}|p{0.8cm}| }
% 	\hline
% 	\multicolumn{10}{|c|}{Effect of Freezing different Numbers of Layers (mAP@0.5 $\uparrow$)} \\
% 	\hline
% 	0 & 3 & 5 & 6 & 7 & 8 & 9 & 12 & 17 & 23  \\
% 	\hline
% 	 \textbf{0.758} & 0.72 & 0.631 & 0.572 & 0.399 & 0.445 & 0.351 & 0.361 & 0.511 & 0.103  \\
% 	\hline
% \end{tabular}
% \caption{\label{tab:detector_freezing_comparison}
%           Effect of freezing different numbers of layers of the network (configuration X) and training for 500 epochs versus no freezing. The first column represents training all layers, the second only the layers up to "functional unit" (FU) 3, the third up to FU 6, and so on. A FU is a set of individual layers, such as a number of convolutional layers. FU 1-9 represents the feature extractor/backbone (CSP-Darknet53), followed by the neck, and the head (FU 13-24). The exact mapping of FU to layers can be found \href{https://github.com/ultralytics/yolov5/blob/master/models/yolov5x.yaml}{here}. Note that we double-checked column "17". Evaluated on real-world test data.}
% \end{table}

\section{Quantitative Results}

% \begin{figure}[tbp]
%   \centering
%     \includegraphics[width=0.8\textwidth]{images/trumpf_test_samples_000.jpg}
%   \caption{\label{fig:test_samples}
%           Samples of our real-hardware and synthetic test data. First two rows show samples from the real data set, row three from our synthetic test data. }
% \end{figure}

% \begin{figure}
%   \centering
%      \includegraphics[width=0.48\textwidth]{yolo_x_results/26d/results.png}
%   \caption{\label{fig:detector_training_results}
%           Box loss, object loss, class loss curves, precision, recall, and mAP during our training. One clearly sees the steep decline during the first, most important epochs. Our model 0.5D (10 epochs) already achieves more than 80\% of the final performance. 4D (80 epochs) is only slightly worse than 26D (500 epochs) in all metrics but precision and recall, which still improves significantly. Most noteably, we do not witness any significant overfitting to our training dataset even close to 500 epochs.}
% \end{figure}
We evaluate our synthetically trained models on both a disjunct set of renderings with unseen object appearances, as well as on real-world photographs. All of our photographs are taken with the HoloLens 2 main camera, the ground truth poses are registered via fiducial markers and OpenCV. Initially, we also wanted to provide a third set of images using the depth sensor of the Microsoft Azure Kinect, but due to the parts being made of sheet metal (specular reflection), registering our models with the objects failed. The benefit of using the HoloLens camera is that the results are representative for the real-world scenario. Our real test set consists of 207 images randomly taken from our video sequence, but all fulfilling the condition that at least 2 objects are visible. For our synthetic test set, we created a metal material and rendered 2500 images with natural object placement using gravity and collision calculation. %We show samples for both in Figure \ref{fig:test_samples}. \\
We test the detection step individually, as well as the pose estimation end-to-end. Also we report the detection results per object class and provide all shapes to show how different geometries perform.

\begin{table*}%[t]
\centering
\caption{\label{fig:ObjectDetectionResults}
          Object detection results (mAP@0.5, higher is better. Best results in \textbf{bold font}.) per object at 3 different numbers of epochs, together with a depiction of the shape, and its dimensions in millimeter. There are clearly more difficult (such as 01, 02, 06, 13, 14) and less difficult objects (08, 09, 11, 15) when measuring difficulty by the speed with which a high-recall is reached per object. Notably, the arguably easier to detect objects are recalled after only a few epochs (10 epochs) and do not improve much, whereas the more difficult objects benefit a lot from continued training. We do not witness any signs of overfitting. Note, that all objects are equally represented in our training data. Evaluated on real-world test data.}
\begin{tabular}{ |p{0.4cm}|p{1.4cm}|p{1.3cm}|p{0.8cm}|p{0.8cm}|p{0.9cm}||p{0.4cm}|p{1.4cm}|p{1.3cm}|p{0.8cm}|p{0.8cm}|p{0.9cm}|}
	\hline
	\multicolumn{12}{|c|}{Detection Results (mAP\@0.5 $\uparrow$)} \\
	\hline
	ID & Shape & Dim. & 10 & 80 & 500 & ID & Shape & Dim. & 10 & 80 & 500\\
	\hline
	01 & \raisebox{-.5\height}{\includegraphics[height=7mm,width=14mm,keepaspectratio]{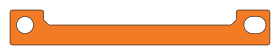}} & 260 x 35 & 0.400 & 0.610 & \textbf{0.709} & 
	09 & \raisebox{-.5\height}{\includegraphics[height=7mm,width=14mm,keepaspectratio]{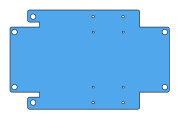}} & 159 x 99 & 0.901 & 0.957 & \textbf{0.980} \\
	02 & \raisebox{-.5\height}{\includegraphics[height=7mm,width=14mm,keepaspectratio]{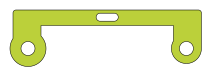}} & 191 x 57 & 0.418 & 0.520 & \textbf{0.584} &
	10 & \raisebox{-.5\height}{\includegraphics[height=7mm,width=14mm,keepaspectratio]{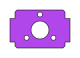}} & 60 x 38 & 0.477 & 0.554 & \textbf{0.627} \\
	03 & \raisebox{-.5\height}{\includegraphics[height=7mm,width=14mm,keepaspectratio]{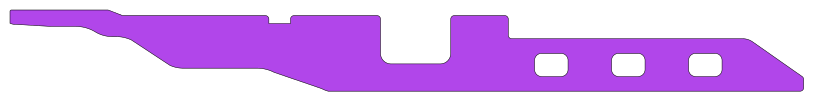}} & 794 x 81 & 0.604 & 0.757 & \textbf{0.854} &
	11 & \raisebox{-.5\height}{\includegraphics[height=7mm,width=14mm,keepaspectratio]{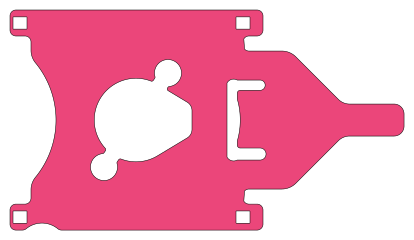}} & 394 x 220 &  0.909 & 0.915 & \textbf{0.937} \\
	04 & \raisebox{-.5\height}{\includegraphics[height=7mm,width=14mm,keepaspectratio]{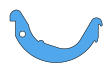}} & 92 x 53 & 0.458 & 0.529 & \textbf{0.676} &
	12 & \raisebox{-.5\height}{\includegraphics[height=7mm,width=14mm,keepaspectratio]{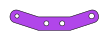}} & 89 x 20 & 0.690 & 0.809 & \textbf{0.875} \\
	05 & \raisebox{-.5\height}{\includegraphics[height=7mm,width=14mm,keepaspectratio]{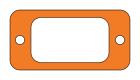}} & 120 x 60 & 0.807 & 0.901 & \textbf{0.934} &
	13 & \raisebox{-.5\height}{\includegraphics[height=7mm,width=14mm,keepaspectratio]{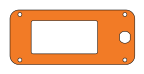}} & 125 x 55 & 0.163 & 0.403 & \textbf{0.513} \\
	06 & \raisebox{-.5\height}{\includegraphics[angle=90,height=7mm,width=14mm,keepaspectratio]{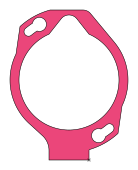}} & 150 x 118 & 0.449 & 0.756 & \textbf{0.779} &
	14 & \raisebox{-.5\height}{\includegraphics[height=7mm,width=14mm,keepaspectratio]{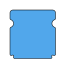}} & 46 x 49 & 0.034 & 0.266 & \textbf{0.378} \\
	07 & \raisebox{-.5\height}{\includegraphics[angle=90,height=7mm,width=14mm,keepaspectratio]{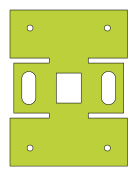}} & 156 x 117 & 0.425 & 0.537 & \textbf{0.638} &
	15 & \raisebox{-.5\height}{\includegraphics[height=7mm,width=14mm,keepaspectratio]{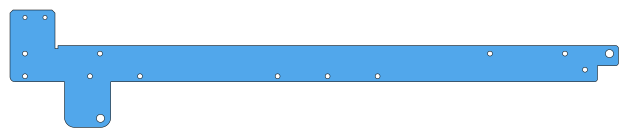}} & 609 x 117 & 0.871 & 0.904 & \textbf{0.948} \\
	08 & \raisebox{-.5\height}{\includegraphics[height=7mm,width=14mm,keepaspectratio]{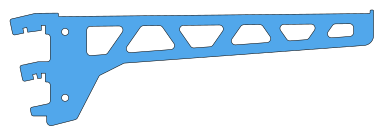}} & 364 x 116 & 0.848 & 0.908 & \textbf{0.933} 
	& & & & & & \\
	\hline
\end{tabular}
\end{table*}

\begin{figure*}[tbp]
  \centering
  \mbox{}
    \includegraphics[width=0.49\textwidth]{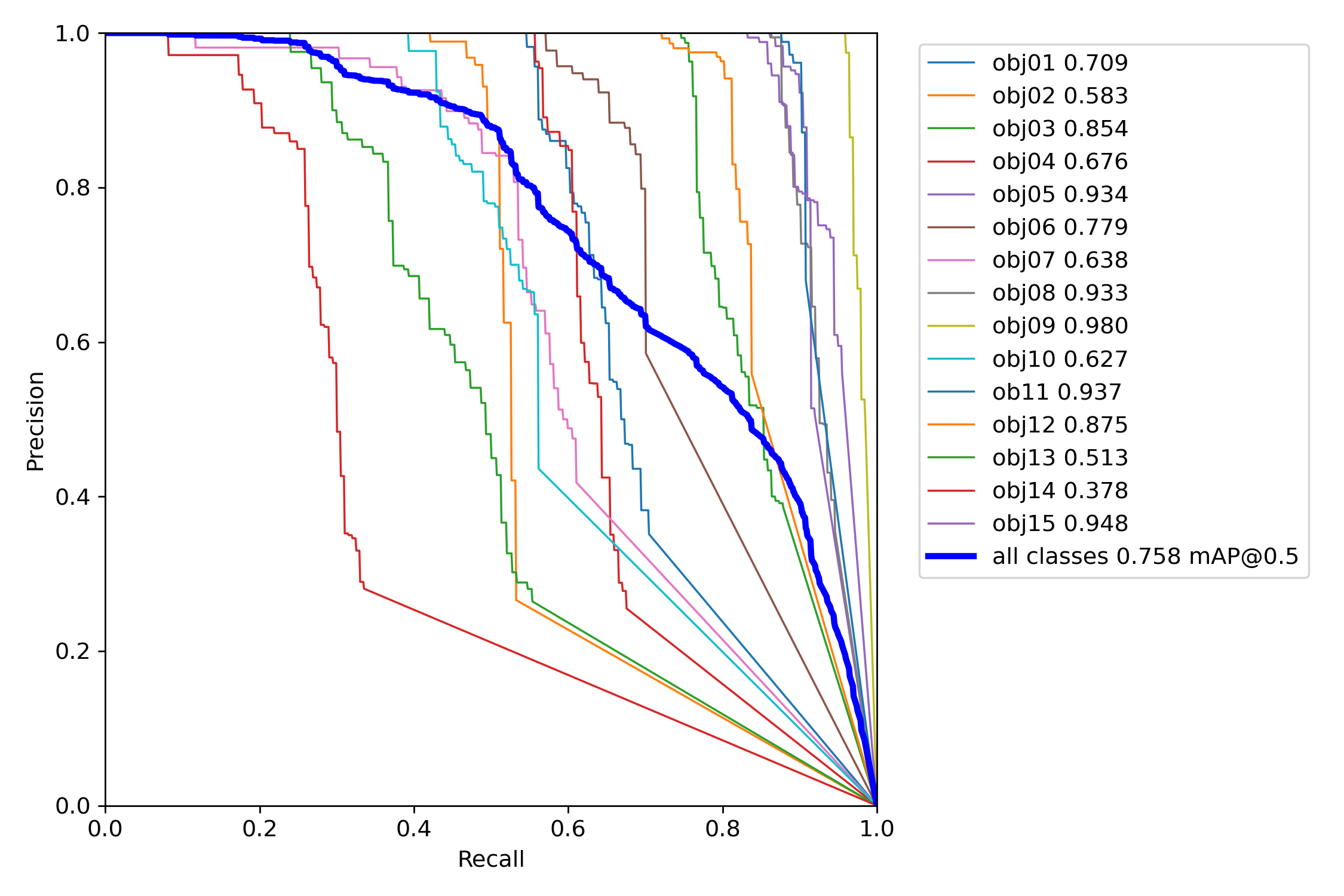}
    \includegraphics[width=0.49\textwidth]{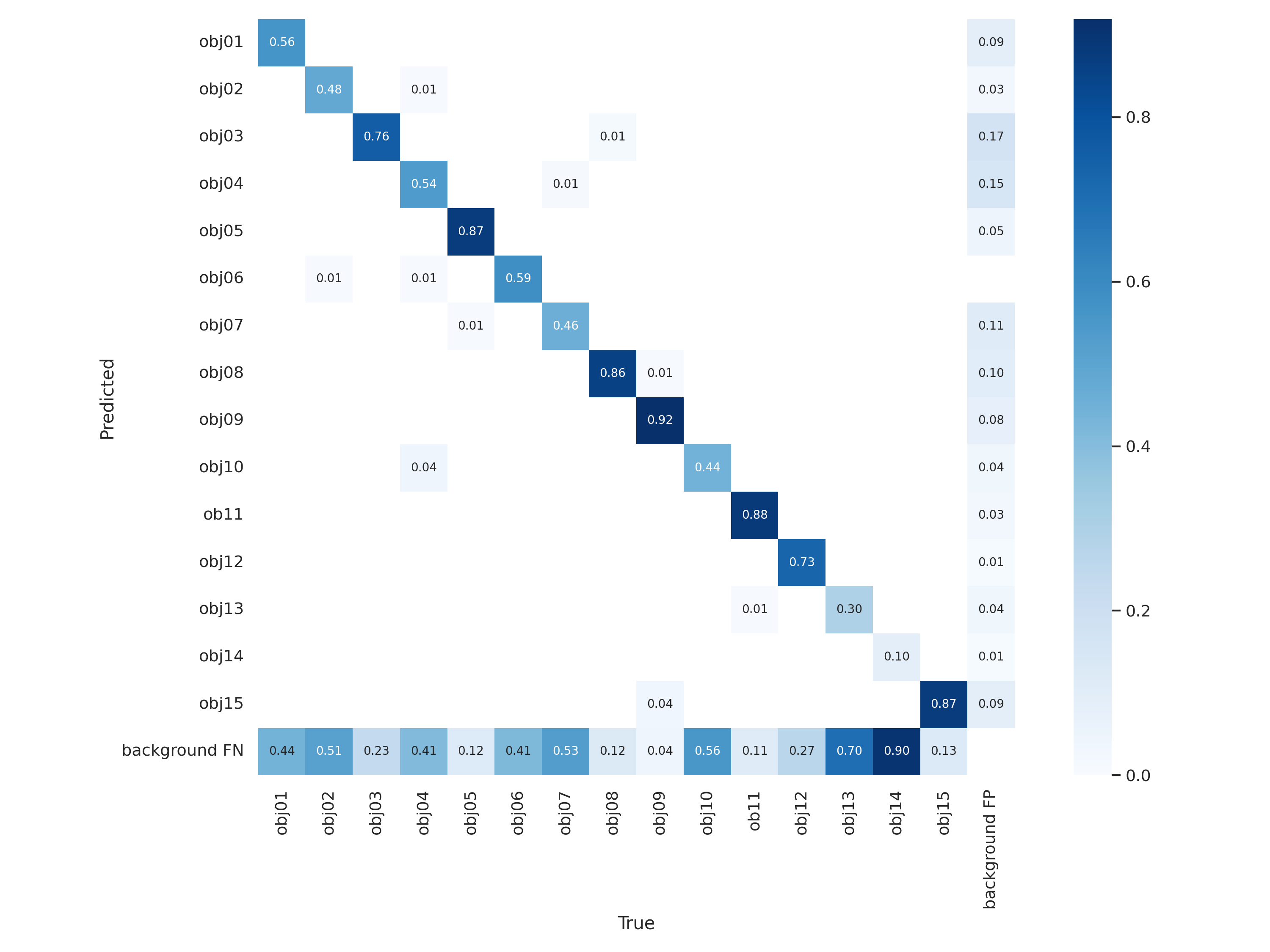}
  \caption{\label{fig:detection_graphs}%
           Precision-recall curves per object and confusion matrices of our objects. As seen in Figure \ref{fig:ObjectDetectionResults} for most objects, the network learns a meaningful representation in only a few epochs, but still improves even close to 500 epochs. Especially the worst performers, such as obj14, improve significantly between epoch 80 and epoch 500. The confusion matrices show that our model usually errs on the side of not detecting an object occurrence (false negative) and much less so on providing erroneous detections (false positives). There are hardly any mix-ups between objects (errors in object class). Evaluated on real-world data. }
\end{figure*}

\subsection{Detection}

\subsubsection{Determining Model Size and Parameters}
We evaluated various model sizes, parameter choices, and dataset sizes and compositions to train our object detector. Performance with various model sizes is presented in Table \ref{tab:detector_model_size_comparison}. Since we train exclusively on renderings of our objects, we experimented with freezing various numbers of layers, following the idea of \cite{pretraining}. Table \ref{tab:detector_freezing_comparison} shows a comparison of the effect of freezing a different number of layers on model size X versus keeping all parameters trainable. However, contrary to the finding of the above paper, our performance drops significantly, even when only freezing the first stage of the feature extractor, and even more drastically when freezing an increasing percentage of the network. Interestingly, when freezing more than just the feature extractor, the performance increases between (9, 12, and 17), before falling off again with freezing more layers (23). We suppose our high quality training data generalizes well to our real world HoloLens 2 images. Given this premise, freezing bigger portions of the network decreases its potential to fit the target objects and thereby leads to reduced performance. As for why freezing more than just the feature extractor can increase performance, compared to freezing fewer layers, is a question we want to look into in future work.  \\
% Figure \ref{fig:detector_training_results} shows the losses on both the training (first row), as well as on the test set (second row). 
We do not use a validation set for early stopping (or to derive any other information for that matter), because of the constraint of not using any real-world data. One could validate on a split of the training data, but we did not notice an improvement in our results, when testing this approach.

\begin{table}%[t]
\centering
\caption{\label{tab:detector_freezing_comparison}
           Effect of freezing different numbers of layers of the network (configuration X) and training for 500 epochs versus no freezing. Columns labeled "Blocks" specify up to which layer block the model has been freezed: 0 means fully trainable, 3 means freezed up to (and including) block 3, and so on. Note that a block represents a set of layers, such as multiple convolutions. Blocks 1-9 represent the feature extractor/backbone (CSP-Darknet53), followed by the neck, and the head (13-24). The exact architecture and mapping of blocks to underlying layers can be found \href{https://github.com/ultralytics/yolov5/blob/master/models/yolov5x.yaml}{here}. Note that we double-checked results for freezing up to (and including) block 17: mAP@0.5 indeed increases, like it did (much more slightly) between 9 and 12. Evaluated on real-world test data. Best result in \textbf{bold font}.}
\begin{tabular}{ |p{1.2cm}|p{1.2cm}|p{1.2cm}|p{1.2cm}|p{1.2cm}|p{1.2cm}| }
	\hline
	\multicolumn{6}{|c|}{Effect of Freezing different Numbers of Layers (mAP@0.5 $\uparrow$)} \\
	\hline
	    \#Blocks & Result & \#Blocks & Result & \#Blocks & Result \\
	    \hline
	    0 & \textbf{0.758}  & 7     & 0.399     & 17    & 0.511 \\
	    3 & 0.72            & 8     & 0.445     & 23    & 0.103 \\
	    5 & 0.631           & 9     & 0.351     &       &  \\
	    6 & 0.572           & 12    & 0.361     &       &  \\
	\hline
\end{tabular}
\end{table}

\subsubsection{Evaluation of our Model}
We trained configuration X of YOLOv5, the largest version, which is still a magnitude faster than the pose estimation step. We compare 3 different models, each consecutive model trained for a longer period of time (10, 80, and 500 epochs), to compare the training time / performance ratio. Results are presented in Table \ref{fig:ObjectDetectionResults}.
We show the precision-recall curves and the confusion matrices for the final model (500 epochs) in Figure \ref{fig:detection_graphs}. The PR curves illustrate the performance differences between different objects, whereas the confusion matrix shows how often an object gets mistaken for another object, not detected (both false negatives) or incorrectly detected as present (false positive).\\
Clearly, object 14 performs the worst of all objects, followed by object 13. We suspect the reason for the bad detection rate of object 14 is its small size compared to the remaining objects. As for object 13, its comparatively low performance comes as a surprise, since the very similar object 05 is among the best performers. We assume this is a random effect, the result of our randomly sampled camera views when creating the ground truth. One also notes that very similar objects, both in shape as well as in size, can perform very differently (object 13 and object 05, or object 01 and 02). As for why this is the case, we do not know, since we draw from an uniform distribution when rendering, giving each and every object the same probability to appear on an image and leading to a relatively equal representation in the training set. To solve this problem we propose to fine-tune the model on a second dataset, which over-represents the difficult objects.

\subsection{Pose Estimation}

\begin{table*}[t]
\centering
\caption{\label{tab:pose_estimation_results}Results of our end-to-end pipeline on our test data. Total number of targets: 2824 (Real) / 31635 (Synth). Threshold is a percentage of the object diameter for MSSD (i.e. 5\% times object diameter), 5r, 10r .. for $r=\frac{image\ width}{640}$ for MSPD, and $\tau=0.15$ for VSD respectively. In all cases, higher is better.}
\begin{tabular}{ |p{2.2cm}||p{1.4cm}|p{1.4cm}|p{1.4cm}|p{1.4cm}|p{1.4cm}|p{1.4cm}| }
	\hline
	\multicolumn{7}{|c|}{End-to-End Pose Estimation Results} \\
	\hline
	THRESHOLD & MSSD $\uparrow$ Synth & MSSD $\uparrow$ Real & MSPD $\uparrow$ Synth & MSPD $\uparrow$ Real & VSD $\uparrow$ Synth & VSD $\uparrow$ Real\\
	\hline
	THR=50 & 0.2622 & 0.0948 & 0.4704 & 0.1866 & 0.292 & 0.0981   \\
    THR=40 & 0.2228 & 0.0894 & 0.4113 & 0.1631 & 0.2685 & 0.0868   \\
    THR=30 & 0.1917 & 0.0824 & 0.339 & 0.1356 & 0.2461 & 0.0705  \\
    THR=20 & 0.1618 & 0.0765 & 0.2781 & 0.1045 & 0.223 & 0.0436   \\
    THR=10 & 0.1148 & 0.0658 & 0.226 & 0.0935 & 0.129 & 0.0103   \\
    THR=05 & 0.0673 & 0.0377 & 0.2074 & 0.0886 & 0.0287 & 0.0014   \\
	\hline
\end{tabular}
\end{table*}

\begin{table}
\centering
\caption{\label{tab:pose_estimation_averages}Average recall across all objects. Total number of targets: 2824 (Real) / 31635 (Synth). Higher is better. We compare our results, both on physically-based rendered images, as well as on our real world sequence, to CosyPose on a dataset with some similarity to ours, as reported at \hyperref[https://bop.felk.cvut.cz/sub_info/2203/]{BOP Challenge}. Note that the ITODD dataset is, unlike ours, recorded with a static camera, avoiding problems such as motion blur. For comparison, we also include CosyPose performance on a strongly textured dataset (YCB-V).}
\begin{tabular}{ |p{5cm}||p{1.2cm}|p{1.2cm}|p{1.2cm}|p{1.2cm}| }
	\hline
	\multicolumn{5}{|c|}{Pose Estimation Averages $\uparrow$} \\
	\hline
	 & Synth & Real & C$_{ITODD}$ & C$_{YCB-V}$ \\
	\hline
	BOP19 AVG RECALL & 0.2295 & 0.0985 & 0.184 & 0.471  \\
	BOP19 AVG RECALL (VSD) & 0.2489 & 0.0573 & - & -  \\
    BOP19 AVG RECALL (MSPD) & 0.3305 & 0.1331 & - & -  \\
    BOP19 AVG RECALL (MSSD) & 0.179 & 0.0812 & - & -  \\
	\hline
\end{tabular}
\end{table}

Since we illustrate the applicability with an assistance application, we are primarily interested in detecting objects sufficiently well. The relevant metrics are Maximum Symmetry-Aware Surface Distance (MSSD) and Maximum Symmetry-Aware Projection Distance (MSPD). MSSD is especially relevant for applications involving robotic manipulation, whereas MSPD ignores the alignment along the view axis, making it highly relevant for AR applications. For comparison with results from the BOPChallenge, we also calculate and report the Visible Surface Discrepancy (VSD) metric. Details for all three metrics can be found in \cite{bopchallenge2020}. We provide our results in Table \ref{tab:pose_estimation_results} and Table \ref{tab:pose_estimation_averages}. 
% \\
% We consider our approach real-time since, by anchoring the predictions we get from the backend in the real world coordinate system (processing roughly 3 frames per second), the user experience is not significantly degraded by head/camera movements. Only when the object is moved with regards to the environment, a pose-update by the NN is needed to superimpose it again (which takes 1/3 of a second).

\subsection{Discussion}
Comparing recall on our photo-realistic renderings with results of Mask-RCNN + CosyPose on similar datasets, such as ITODD, we achieve noticeably higher performance (Table \ref{tab:pose_estimation_averages}). Once we use our real-hardware test sequence, however, performance is almost cut in half. We hypothesize at least some of the reason for this has to do with us using the HoloLens 2 camera for ground truth creation. Although the camera uses a nominally high resolution, when looking at individual frames, they are often extremely blurry, especially during movement. We still wanted to use the use-case relevant, head-mounted camera system, because we are interested in real-world performance. Also, the ITODD dataset is recorded with a static camera, avoiding our problem of vanishing detections with certain camera movements. When comparing to one of the best performing datasets (as presented by the BOPChallenge 2020) YCB-V, we argue that most of the difference is due to YCB-V being strongly texturized, since texture is a strong cue readily-exploited by CNNs \cite{texture_overfitting}. As shown above, we do not make strong assumptions about the materials used in our training images, randomizing texture as well as reflection behavior. \\
As for our per object detection results, we find a generally strong performance measured in recall. We compared several model sizes and found that even the biggest models improve the result on our test set, without overfitting. Also, we showed that freezing certain layers of the feature extractor did, contrary to previous research, degrade results. As for the problem of camera movements leading to few or no detections for certain frames, one would focus on adding samples with this characteristic to the training data. We already noted that one probably needs to specifically address the more difficult to detect objects to further improve detection performance. \\
Concerning the user experience, because of our sending and receiving data to and from the backend, we experience noticeable latency. While we do think this latency does not hinder the application, reducing it would certainly improve the experience.

\section{Conclusion and Future Work}
\label{future_work}
We presented an easily adjustable end-to-end pipeline for detection and pose estimation of flat, texture-less industrial objects only requiring vector graphic representations for training, and evaluated on an edge devices (HoloLens 2). These results are reproducible without extensive expertise in the application domain, in order to, for example, manually create the scenes for training data acquisition. \\
Future work could focus on improving inference speed, in order to run the solution natively on low power edge devices. This would drastically reduce latency and improve the user experience.
Another improvement, one we are currently evaluating, is the inclusion of a tracker running on the frontend device. This still requires communicating with a backend for processing, but also does away with the latency. 
Also, a user study could compare speed and accuracy at performing the sorting task, providing the manufacturing document on paper, using the iPad application, and the hands-free HoloLens 2 application.
Finally, our experiments with freezing parts of the detection network motivate further investigation. \newline

% \textbf{Note on reproducibility.} For copyright reasons we cannot share our object models, nor the manufactoring documents. Our objects (or very similar ones), however, can easily be reproduced with our provided information. The same applies to our pipeline, since Blender, YOLOv5, and CosyPose are free and/or open source and we shared the most significant procedures and parameterizations. 

% blind
\textbf{Acknowledgment}: This research was funded in parts by the German Federal Ministry of Education and Research. \\
\bibliographystyle{splncs04}
\bibliography{bibliography}

\end{document}